%% file: main.tex
\PassOptionsToPackage{unicode}{hyperref}
\PassOptionsToPackage{hyphens}{url}
\documentclass[10pt]{article}
\usepackage{xcolor}
\usepackage{amsmath,amssymb}
\usepackage[a4paper,margin=0.82in]{geometry}
\usepackage[section]{placeins}
\usepackage{flafter}
\usepackage{caption}
\usepackage{fancyhdr}
\usepackage{iftex}
\ifPDFTeX
  \usepackage[T1]{fontenc}
  \usepackage[utf8]{inputenc}
  \usepackage{textcomp}
\else
  \usepackage{unicode-math}
  \defaultfontfeatures{Scale=MatchLowercase}
  \defaultfontfeatures[\rmfamily]{Ligatures=TeX,Scale=1}
\fi
\usepackage{lmodern}
\IfFileExists{upquote.sty}{\usepackage{upquote}}{}
\IfFileExists{microtype.sty}{
  \usepackage[]{microtype}
  \UseMicrotypeSet[protrusion]{basicmath}
}{}
\makeatletter
\@ifundefined{KOMAClassName}{
  \IfFileExists{parskip.sty}{
    \usepackage{parskip}
  }{
    \setlength{\parindent}{0pt}
    \setlength{\parskip}{6pt plus 2pt minus 1pt}}
}{\KOMAoptions{parskip=half}}
\makeatother
\usepackage{longtable,booktabs,array}
\usepackage{calc}
\usepackage{etoolbox}
\makeatletter
\patchcmd\longtable{\par}{\if@noskipsec\mbox{}\fi\par}{}{}
\makeatother
\IfFileExists{footnotehyper.sty}{\usepackage{footnotehyper}}{\usepackage{footnote}}
\makesavenoteenv{longtable}
\usepackage{graphicx}
\usepackage{float}
\makeatletter
\newsavebox\pandoc@box
\newcommand*\pandocbounded[1]{
  \sbox\pandoc@box{#1}
  \Gscale@div\@tempa{\textheight}{\dimexpr\ht\pandoc@box+\dp\pandoc@box\relax}
  \Gscale@div\@tempb{\linewidth}{\wd\pandoc@box}
  \ifdim\@tempb\p@<\@tempa\p@\let\@tempa\@tempb\fi
  \ifdim\@tempa\p@<\p@\scalebox{\@tempa}{\usebox\pandoc@box}
  \else\usebox\pandoc@box
  \fi
}
\def\fps@figure{htbp}
\makeatother
\providecommand{\tightlist}{\setlength{\itemsep}{0pt}\setlength{\parskip}{0pt}}
\usepackage[]{natbib}
\usepackage{bookmark}
\IfFileExists{xurl.sty}{\usepackage{xurl}}{}
\hypersetup{
  pdftitle={Do Visual Grounding Decoders Need Feed-Forward Networks?: A Controlled Study over Frozen Vision-Language Features},
  pdfauthor={Tarun Tomar},
  colorlinks=true,
  linkcolor=black,
  citecolor=blue!45!black,
  urlcolor=blue!45!black,
  pdfcreator={LaTeX}}

\title{Do Visual Grounding Decoders Need Feed-Forward Networks?}
\usepackage{etoolbox}
\makeatletter
\providecommand{\subtitle}[1]{\apptocmd{\@title}{\par {\large #1 \par}}{}{}}
\makeatother
\subtitle{A Controlled Study over Frozen Vision-Language Features}
\author{Tarun Tomar\\\small\url{https://github.com/TarunTomar122/attention-is-all-you-need-for-vlms}}
\date{}

\begin{document}
\maketitle

\begin{abstract}
Do feed-forward networks (FFNs) in visual grounding decoders add essential
computation once a pretrained vision-language model has already encoded image
and language context? We compare a four-block attention-only decoder (A4), a
matched four-block attention-plus-FFN decoder (S4), and an eight-block
attention-only parameter control (A8) over frozen VLM features. A4 matches or
slightly exceeds S4 on RefCOCOg and Ref-Adv-s. FineCops-Ref reveals a small A4
deficit of 0.52 percentage points at IoU@0.5 (95\% CI [0.12, 0.95] in favor of
S4), but A8 recovers it and finishes 0.26 points above S4. Official FineCops
levels do not show a monotonic increase in the gap. A4 reduces trainable
decoder parameters by 44.4\% and cached-decoder latency by 10.1\%, although
end-to-end latency remains backbone-dominated. These results concern the
trainable grounding decoder, not a complete attention-only VLM.
\end{abstract}

\section{Introduction}\label{introduction}

Visual grounding maps an image and a referring expression to one target box.
Modern grounding systems commonly use Transformer decoders that combine
self-attention or cross-attention with token-wise feed-forward networks. This
design is inherited from the standard Transformer block and remains common in
systems that fuse language and image features, including MDETR and Grounding
DINO \citep{kamath2021mdetr,liu2024groundingdino}. The complete block is a
reasonable default when a model must build its representations from scratch.
It is less clear whether every sublayer remains necessary in a small task head
whose inputs have already been contextualized by a large pretrained model.

This paper isolates that narrower question. A frozen vision-language model
supplies spatial image tokens and contextual text states. A small trainable
decoder then retrieves the region described by the expression and produces a
patch heatmap. We remove only the FFN residuals inside that decoder while
holding its frozen features, learned projections, attention layers, readout,
loss, optimizer, box conversion, and evaluation protocol fixed. The frozen
backbone retains its pretrained FFNs throughout.

The comparison matters because attention and FFNs provide different inductive
biases. Cross-attention routes a query toward relevant context. An FFN applies
the same token-wise nonlinear map after that routing step. If the frozen
features already expose object identity, attributes, spatial location, and
language context, the decoder may mostly need retrieval and assembly. If the
decoder must construct new relational features, the FFN may remain useful even
when attention can localize individual tokens.

We study three questions:

\begin{enumerate}
\def\labelenumi{\arabic{enumi}.}
\tightlist
\item At fixed decoder depth, does A4 retain the grounding performance of S4?
\item If a gap appears, does it grow with released or official difficulty metadata?
\item Can additional attention depth recover a gap after the FFN parameter budget is reallocated?
\end{enumerate}

The answer is mixed but coherent. On RefCOCOg direct expressions, A4 is 0.26
percentage points above S4 and satisfies the predeclared retention gate, while
the planned direct-versus-logical interaction is not confirmed. On Ref-Adv-s,
A4 is 0.79 points above S4 and does not degrade monotonically with negation,
expression length, or distractor count. FineCops-Ref exposes the only clear
fixed-depth deficit: A4 is 0.52 points below S4. Doubling attention-only depth
to A8 recovers that overall gap. These results reject both simple extremes:
FFNs are neither universally essential nor uniformly irrelevant in the tested
decoder.

\begin{figure}[t]
\centering
\includegraphics[width=0.98\linewidth]{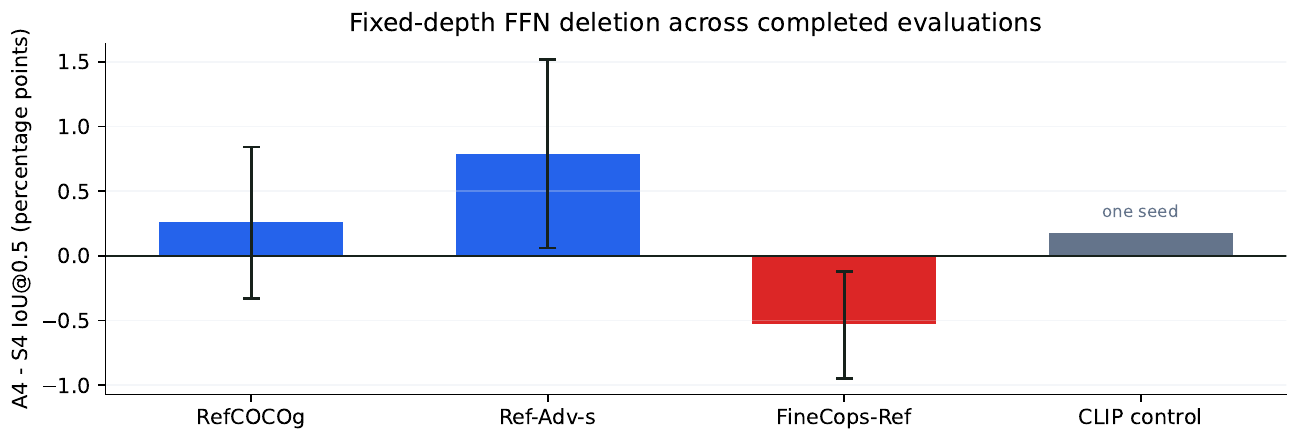}
\caption{The fixed-depth A4-S4 comparison across completed evaluations. Error bars show the prespecified interval where available; the CLIP control is one-seed descriptive evidence. Positive values favor A4.}
\label{fig:evidence-overview}
\end{figure}

Our contributions are:

\begin{enumerate}
\def\labelenumi{\arabic{enumi}.}
\tightlist
\item A causal FFN-deletion study over identical frozen image and text features, with a parameter-matched attention-depth control.
\item A difficulty-oriented evaluation across standard, adversarial, and controlled compositional grounding, supported by modality shuffles and paired image-clustered intervals.
\item An evidence-first release that preserves mixed and negative outcomes, including the unconfirmed logical interaction and the absence of a monotonic difficulty boundary.
\end{enumerate}

\section{Motivation}\label{motivation}

\subsection{Does a frozen representation change what the decoder needs?}\label{does-a-frozen-representation-change-what-the-decoder-needs}

The standard Transformer alternates attention with a position-wise FFN
\citep{vaswani2017attention}. Attention moves information between tokens,
whereas the FFN transforms each token independently. In a large encoder or
generative model, the two operations jointly build and refine representations
across many layers. A grounding decoder over frozen VLM tokens operates under a
different contract. Its image input already contains a dense raster of
contextual features, and its text input already encodes the complete referring
expression. The remaining output is one spatial selection.

Prior work establishes that pretrained attention contains useful localization
signals. Kang et al. identify a small set of grounding-relevant attention heads
inside a frozen LVLM \citep{kang2025heads}. F-LMM translates frozen multimodal
attention maps into segmentation masks with CNN and SAM refinement
\citep{wu2025flmm}. ReCLIP shows that frozen contrastive representations can
score candidate regions, although spatial relations remain difficult without
an added resolver \citep{subramanian2022reclip}. These results make an
attention-only decoder plausible, but none isolates whether a newly trained,
matched grounding head needs its token-wise FFNs.

Our single-query design sharpens the question. Query self-attention would be
degenerate because there is only one query, so each block alternates text and
image cross-attention. If the query can retrieve the relevant linguistic and
visual features directly, an FFN may add parameters without changing the
localization decision. Conversely, compositional expressions may require a
nonlinear transformation after retrieval. The experiment therefore needs
controlled difficulty evidence rather than one aggregate score.

\subsection{Why an overall match is insufficient}\label{why-an-overall-match-is-insufficient}

A4 could match S4 for several uninteresting reasons. A dataset might contain
many direct noun references and too few relational expressions to expose a
difference. A learned position prior could produce acceptable boxes without
using language. A model could rely on image salience and ignore the expression,
or use text length as a shortcut. A heatmap threshold tuned after test access
could hide representation differences. Any of these cases would make the
architectural conclusion unreliable.

We address these alternatives with a frozen selection boundary and three kinds
of controls. First, RefCOCOg includes a predeclared direct-retention test and a
direct-minus-logical interaction. Second, image and text shuffles test whether
both modalities influence the prediction. Third, Ref-Adv-s and FineCops-Ref
provide released metadata for adversarial and compositional difficulty. The
study uses only native or official fields for these slices, avoiding a
post-hoc taxonomy designed around observed errors.

Uncertainty also matters. Several expressions may describe the same image, so
treating all rows as independent would make intervals too narrow. We average
paired model seeds per example and bootstrap image IDs, retaining all
expressions for every sampled image. A slice is interpreted as a boundary only
when its trend and interval support that reading. One noisy level or tuple type
does not become a reasoning claim.

\subsection{Why capacity reallocation needs a separate control}\label{why-capacity-reallocation-needs-a-separate-control}

Removing four FFNs changes both the computation type and the trainable
parameter count. At fixed depth, A4 contains 2.64 million trainable parameters
while S4 contains 4.74 million. If A4 loses accuracy, that result alone cannot
distinguish an FFN-specific advantage from reduced capacity. A matched study
therefore needs a model that spends approximately the same parameter budget on
attention instead.

A8 repeats the attention-only block eight times and contains 4.75 million
trainable parameters, within 0.2\% of S4. This control is parameter-matched,
not compute-matched: it uses more decoder MACs and higher cached-feature
latency. Its purpose is interpretive. If A4 trails S4 and A8 recovers, the
evidence supports attention capacity as a substitute at the measured scale. If
both A4 and A8 trail S4, the result would be more consistent with an advantage
specific to FFN computation.

This distinction follows broader attention-only work. Controlled language-model
experiments report that attention-only models can close much of the gap after
capacity is reallocated into depth \citep{ndubuaku2026attention}. Synthetic
grounded composition work also shows that shallow attention can solve attribute
composition, with additional depth needed for harder relations
\citep{sikarwar2022groundcompose}. Neither result establishes what happens for
natural-image grounding over frozen VLM representations, which motivates the
present matched decoder study.

\section{Methodology}\label{methodology}

\subsection{Frozen features and decoder variants}\label{frozen-features-and-decoder-variants}

The primary backbone is frozen SigLIP2 Base/16 at 384-pixel resolution
\citep{tschannen2025siglip2}. For each image-expression pair it returns 576
image tokens arranged in a $24\times24$ raster and padded contextual text
states. Separate learned linear maps project both streams to decoder width
$d=256$. These projections are shared in design and initialization across
paired variants and are included in trainable-parameter counts.

The checkpoint is \texttt{google/siglip2-base-patch16-384} at pinned revision
\nolinkurl{f775b65a79762255128c981547af89addcfe0f88}. The complete image and text
encoders run in evaluation mode under \texttt{torch.no\_grad()}. Final-layer
spatial image states remain in raster order, and text padding is preserved as
an explicit attention mask. No backbone parameter is fine-tuned, prompt-tuned,
LoRA-tuned, or selectively unfrozen. Feature caching is an execution
optimization only: cached tensors are required to match uncached tensors before
they are used for decoder training or latency isolation.

Images are converted to RGB and resized directly to the backbone's declared
square resolution without cropping. Cropping would be unsafe for grounding
because it can remove or truncate the referred object. Ground-truth coordinates
are scaled independently along the two axes and the inverse factors are stored
for evaluation. The CLIP replication uses the same $24\times24$ grid at
336-pixel resolution by disabling its default center crop. This keeps decoder
token count and output geometry fixed across the two backbone families.

The decoder carries one learned query $q\in\mathbb{R}^{1\times d}$. Each
pre-normalized block first attends to the text tokens $T$ and then the image
tokens $I$:

\[
q \leftarrow q + \operatorname{CrossAttentionText}(\operatorname{LN}(q),\operatorname{LN}(T)),
\]
\[
q \leftarrow q + \operatorname{CrossAttentionImage}(\operatorname{LN}(q),\operatorname{LN}(I)).
\]

S4 then applies

\[
q \leftarrow q + \operatorname{FFN}(\operatorname{LN}(q)),
\]

where the FFN has hidden width 1024 and GELU activation. A4 omits this residual
in each of four blocks. A8 uses eight attention-only blocks. Every variant uses
eight attention heads, zero dropout, learned affine LayerNorm, residual
connections around every retained sublayer, and non-affine per-head query/key
RMS normalization.

\begin{figure}[t]
\centering
\includegraphics[width=0.98\linewidth]{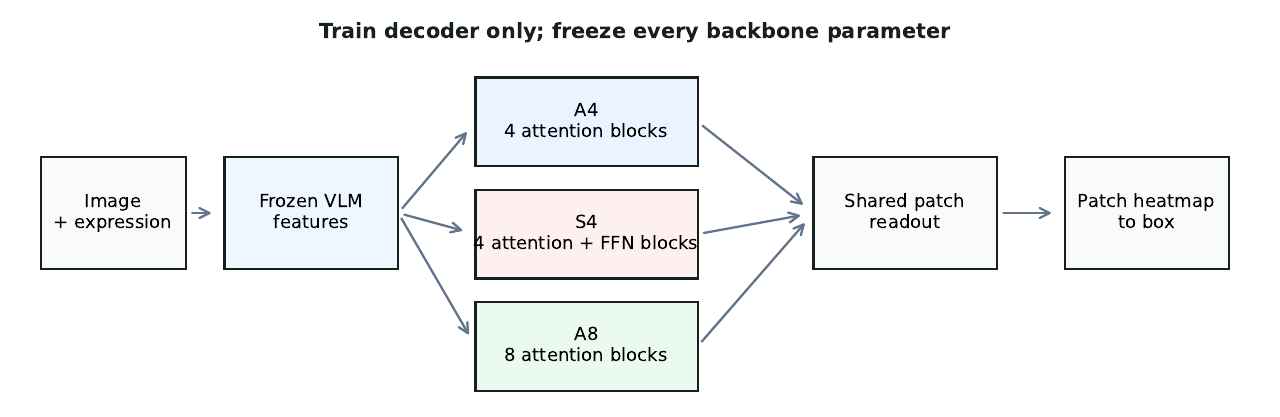}
\caption{Controlled decoder intervention. The VLM remains frozen and the input projections, patch readout, supervision, and box conversion are shared. A4-S4 isolates FFN deletion; A8 tests parameter reallocation into attention depth.}
\label{fig:method-overview}
\end{figure}

\subsection{Shared patch-attention readout}\label{shared-patch-attention-readout}

The prediction head is deliberately separate from the last decoder block. If
the model read out the final block's image-attention weights directly, S4's
last FFN would occur after the prediction and could not affect the loss. The
shared readout instead computes a new compatibility score between the final
query and every projected image token.

For head $h$ and image patch $i$,

\[
s_{h,i}=\frac{\langle \operatorname{RMSNorm}(W^q_hq),
\operatorname{RMSNorm}(W^k_hI_i)\rangle}{\sqrt{32}}.
\]

The readout averages logits across heads and normalizes over all 576 patches:

\[
p_i=\operatorname{softmax}_i\left(\frac{1}{H}\sum_{h=1}^{H}s_{h,i}\right).
\]

The distribution $p$ is the model's only localization output. There is no
coordinate-regression MLP, segmentation refiner, CNN, auxiliary language loss,
multi-query detector, or backbone adaptation. This keeps the causal path small
and prevents an unmeasured nonlinear head from replacing the removed decoder
FFNs.

\subsection{Target distribution and box conversion}\label{target-distribution-and-box-conversion}

Each ground-truth box is converted to a $24\times24$ patch target. For patch
$i$, let $a_i$ be the intersection area between the patch cell and the target
box. The normalized target is

\[
y_i=\frac{a_i}{\sum_{j=1}^{576}a_j},
\]

and training minimizes distributional cross-entropy

\[
\mathcal{L}=-\sum_{i=1}^{576}y_i\log p_i.
\]

At evaluation time, horizontal and vertical marginals are computed from $p$.
For each axis, the decoder selects the shortest contiguous patch interval that
contains at least probability mass $\tau$. Ties prefer greater contained mass
and then the lower start index. Patch edges are mapped back through the recorded
resize and crop geometry to the original image.

The mass threshold was selected once from
$\{0.5,0.6,0.7,0.8,0.9\}$ using S4 on RefCOCOg validation. The selected value,
$\tau=0.8$, was then frozen for every variant, dataset, backbone, and test
split. Heatmap metrics remain visible so the conclusion does not depend only on
one deterministic box conversion.

\subsection{Matched comparisons and controls}\label{matched-comparisons-and-controls}

The causal fixed-depth comparison is A4 versus S4. Both models receive
identical cached frozen features and masks, use the same training budget, and
produce the same output shape. A8 is the parameter control. A zero-block
decoder (D0), a uniform heatmap, and a fixed position prior provide lower-level
diagnostics where available. Image-shuffle and text-shuffle evaluations replace
one modality's pairing while preserving all other code and weights.

The implementation checks that output distributions are finite and sum to one,
padding tokens receive zero text-attention probability, ground-truth patch
distributions are valid, and predicted boxes remain within image bounds. It
also checks that A4 contains no FFN or two-layer token-wise projection and that
S4's final FFN receives gradient from the localization loss. These invariants
ensure that the named ablation matches the executed graph.

Paired variants also share ordered example manifests and immutable output
directories. Preparation fails on duplicate identifiers, missing images,
non-finite coordinates, non-positive boxes, or boxes fully outside an image.
Malformed examples are not silently discarded. Raw predictions are stored
before aggregate metrics are calculated, which keeps thresholded accuracy,
continuous IoU, pointing, target mass, and slice analyses traceable to the same
per-example outputs.

\subsection{Evaluation and uncertainty}\label{evaluation-and-uncertainty}

The primary metric is box accuracy at IoU$\geq0.5$. Mean IoU, pointing
accuracy, and target mass are secondary. Three-seed comparisons average paired
seed outputs per example, then use 10,000 paired percentile bootstrap
replicates with fixed seed 20260812. Image IDs, rather than expression rows,
are resampled so that all references to the same image move together.

For a slice $g$, define the fixed-depth difference in percentage points as

\[
\Delta_g=100\left(A_g^{\mathrm{A4}}-A_g^{\mathrm{S4}}\right).
\]

RefCOCOg direct retention uses its prespecified 90\% interval and practical
margin. All other reported intervals are 95\% descriptive or confirmatory
intervals as recorded by the frozen evaluation contract. Ref-Adv-s bins use
native metadata quartiles fixed before performance slicing. FineCops uses only
official level and tuple-type fields. No LLM-derived semantic categories are
introduced after observing results.

\section{Experiments}\label{experiments}

\subsection{Models, datasets, and metrics}\label{models-datasets-and-metrics}

\textbf{Backbones.} The primary experiments use frozen SigLIP2 features. A
one-seed frozen CLIP-L/14@336 experiment checks whether the direction is unique
to the primary backbone. This control preserves the decoder definitions,
training budget, loss, and frozen mass threshold, but it is descriptive rather
than a second multi-seed confirmation.

\textbf{Data.} RefCOCOg UMD is the primary three-seed evaluation and supplies
the direct/logical protocol \citep{yu2016context}. RefCOCO UNC provides a
seed-0 classic-dataset replication. Ref-Adv-s contains 1,142 prepared
evaluation-only rows and native negation, length, and distractor metadata
\citep{dong2026refadv}. FineCops-Ref contributes 9,605 official positive-test
examples with released difficulty levels and tuple types
\citep{liu2024finecops}. No Ref-Adv or FineCops example selects a checkpoint,
threshold, architecture, or plot rule.

RefCOCOg contains 80,512 training expressions, 4,896 validation expressions,
and 9,602 official UMD test expressions. The classic annotations link each
sentence to a COCO object box; multiple expressions may describe one object or
image, which motivates image-clustered uncertainty. The RefCOCO seed-0
replication uses its official UNC split and fixed 5,000-update training budget.
Ref-Adv-s is the pinned public subset: its packaged split is renamed to test
inside this project, and its native caption, box, negation, distractor, image
source, and authorship fields are preserved without semantic relabeling.

FineCops evaluation uses only the official positive test examples because the
decoder contract predicts one target box for a valid referring expression. Its
released level and tuple-type annotations are read directly. All third-party
images remain outside Git, and stable IDs plus aggregate or per-example metric
tables are released where redistribution permits. This separates reproducible
evaluation evidence from upstream image licenses.

\textbf{Metrics.} IoU@0.5 is primary because it captures whether the single
predicted box meets the standard localization threshold. Mean IoU retains
continuous overlap information, pointing accuracy measures whether the
highest-probability patch lies inside the target, and target mass measures the
probability assigned to the ground-truth region. Decoder parameters, analytical
MACs, peak memory, and repeated synchronized latency measurements describe
efficiency.

The primary training and evaluation matrix is intentionally asymmetric. The
full three-seed comparison is reserved for the frozen SigLIP2 RefCOCOg models
and the locked Ref-Adv-s and FineCops evaluations. RefCOCO and CLIP answer
narrow replication questions and are labeled seed-0 or one-seed wherever they
appear. This prevents a small descriptive run from being presented with the
same evidential weight as the paired multi-seed results.

\subsection{Does fixed-depth FFN deletion preserve standard grounding?}\label{does-fixed-depth-ffn-deletion-preserve-standard-grounding}

On RefCOCOg direct expressions, A4-S4 is +0.26 percentage points with a 90\%
interval of [-0.33, +0.84]. The lower bound remains above the predeclared
-5-point practical margin, so direct retention passes. On logical expressions,
A4-S4 is -1.04 points with a 95\% interval of [-2.20, +0.15]. The planned
direct-minus-logical interaction is +1.30 points with a 95\% interval of
[-0.09, +2.66], so the stronger retrieval-versus-reasoning boundary is not
confirmed.

The modality controls show that this match does not come from ignoring one
input. Correct pairing exceeds image shuffle by 48.7 points for A4 and 48.6
points for S4. Correct pairing exceeds text shuffle by 22.6 and 22.4 points,
respectively. Both architectures therefore use image and language information
under the tested pipeline. The result supports fixed-depth retention on the
primary dataset while preserving the failed interaction as a negative outcome.

The predeclared interaction was deliberately conjunctive: practical retention
on direct references and a sufficiently large direct-minus-logical difference
were both required. Passing retention alone does not permit the paper to claim
that FFNs become necessary for logical language. This gate matters because the
point estimate moves in the predicted direction, but its uncertainty still
includes zero. Keeping the failed condition visible avoids turning a plausible
post-hoc story into a confirmed result.

\begin{figure}[t]
\centering
\includegraphics[width=0.98\linewidth]{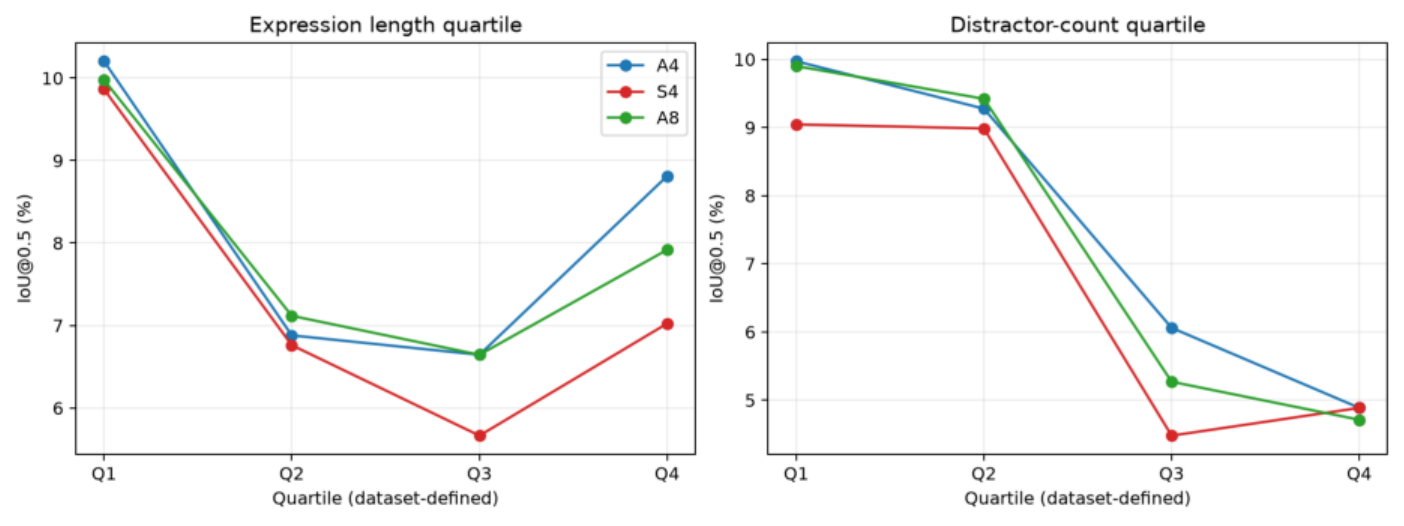}
\caption{Absolute Ref-Adv-s performance across native difficulty metadata. Absolute values remain low for all variants, so the architectural result is a paired comparison rather than a claim of strong adversarial grounding.}
\label{fig:refadv-performance}
\end{figure}

\subsection{Does adversarial difficulty expose an FFN advantage?}\label{does-adversarial-difficulty-expose-an-ffn-advantage}

Ref-Adv-s is evaluation-only and was accessed after model and threshold choices
were frozen. Overall IoU@0.5 is 8.11\% for A4, 7.33\% for S4, and 7.91\% for
A8. The paired A4-S4 difference is +0.79 points with a 95\% interval of
[+0.06, +1.52]. The absolute accuracies show that this is a difficult setting
for the entire decoder family. The comparison should therefore be read as an
architectural stress test, not as a competitive Ref-Adv system.

Negation does not expose a clear A4 loss: the negation slice is -0.15 points
with an interval crossing zero, while the non-negation slice favors A4 by 1.41
points. Expression-length quartiles move from +0.34 to +1.79 points without a
negative trend. Distractor quartiles range from +0.29 to +1.58 points before
ending at 0.00 in the highest-distractor quartile. The longest expressions and
largest distractor sets therefore do not produce the hypothesized monotonic
attention-only collapse.

\begin{figure}[t]
\centering
\includegraphics[width=0.98\linewidth]{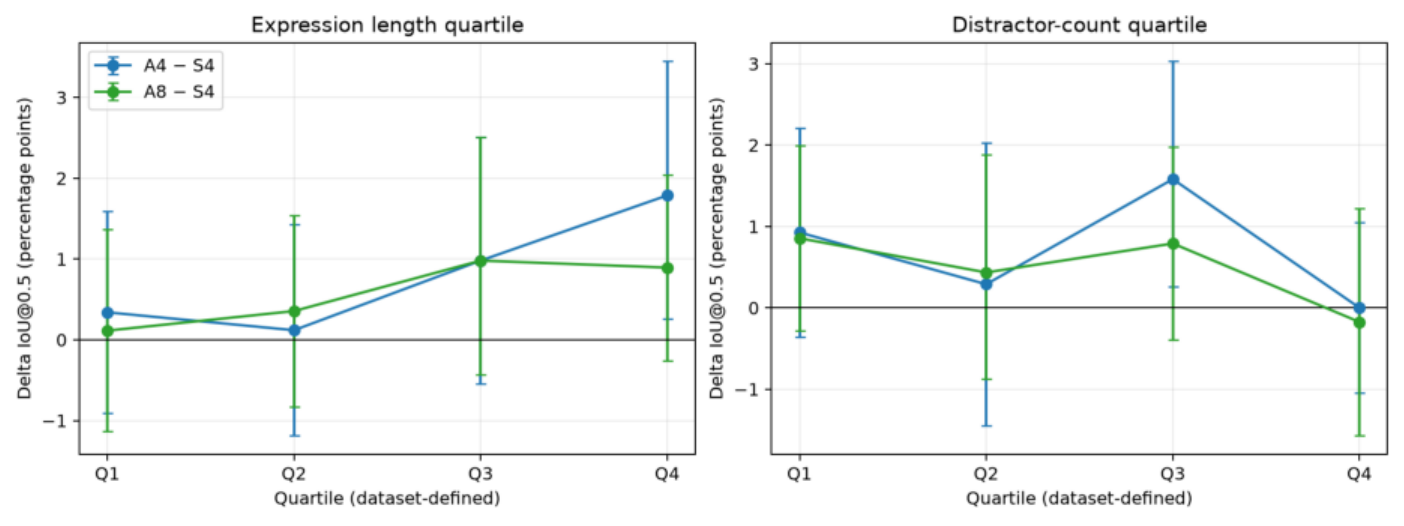}
\caption{Paired A4-S4 and A8-S4 differences on Ref-Adv-s. Native length and distractor slices do not reveal a monotonic fixed-depth deficit.}
\label{fig:refadv-deltas}
\end{figure}

This result narrows the interpretation of difficulty. Released metadata can
identify useful stress slices, but longer language and more distractors are not
equivalent to a controlled requirement for token-wise nonlinear computation.
Ref-Adv may also expose limitations in the frozen representation, 24-by-24
readout, or training distribution that affect A4 and S4 together.

The secondary metrics reinforce the bounded reading. A4 and A8 both reach
approximately 27\% pointing accuracy, and mean IoU differs by less than 0.002
from S4. Thus the positive IoU@0.5 delta is not accompanied by a large shift in
continuous overlap. The adversarial set mainly establishes that FFN deletion
does not uniquely cause the already-low performance regime; it does not show
that any of the three heads solves adversarial reference resolution.

\subsection{Does controlled composition reveal a fixed-depth gap?}\label{does-controlled-composition-reveal-a-fixed-depth-gap}

FineCops-Ref provides the clearest evidence for S4. Across 9,605 official
positive-test examples, A4 reaches 27.01\% IoU@0.5 and S4 reaches 27.54\%.
The paired difference is -0.52 percentage points with a 95\% interval of
[-0.95, -0.12]. This is small in absolute size but is the only completed
three-seed comparison whose interval lies fully below zero for A4 overall.

The official difficulty levels do not turn that overall deficit into a
monotonic boundary. Level 1 is -0.58 points and level 2 is -0.53 points. Level
3 is +0.14 points with a wide interval of [-1.10, +2.55] over 471 examples.
The hardest released level therefore does not show a larger A4 loss. It is too
imprecise to establish either an A4 advantage or equality.

Continuous metrics show how small the architectural difference remains. Mean
IoU is 0.2971 for A4, 0.2978 for S4, and 0.2977 for A8. Pointing accuracy is
52.20\%, 51.79\%, and 52.04\%, respectively, while target mass is 0.3609,
0.3597, and 0.3621. S4's thresholded advantage therefore does not coincide
with a broad advantage on every localization statistic. The paper nevertheless
retains the IoU@0.5 result because it is the frozen primary outcome.

\begin{figure}[t]
\centering
\includegraphics[width=0.98\linewidth]{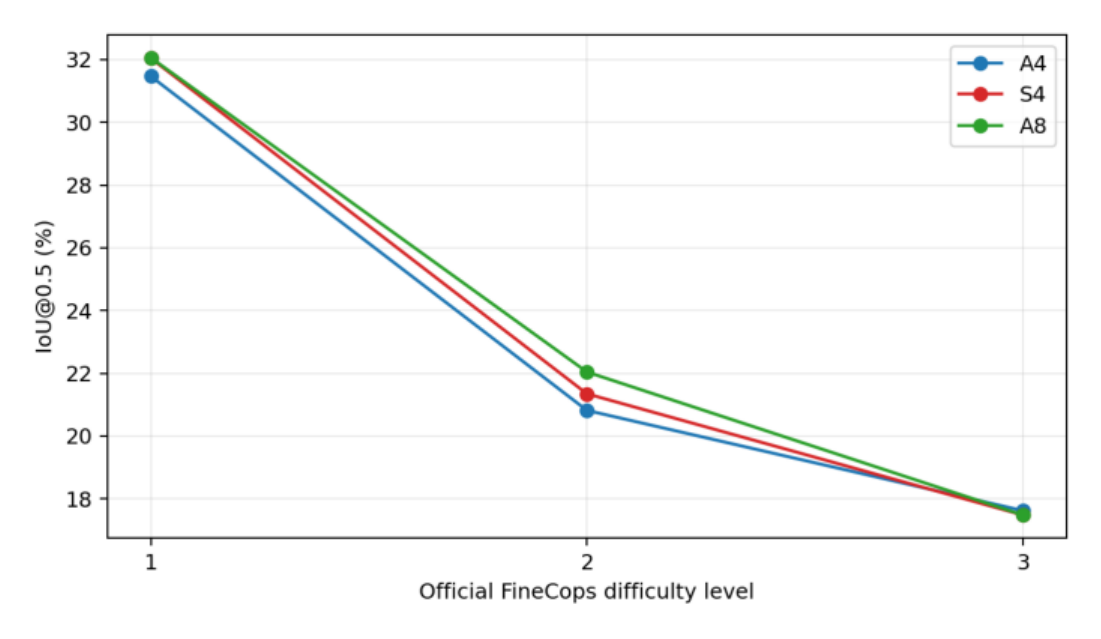}
\caption{Absolute IoU@0.5 across official FineCops-Ref levels. Performance declines with released difficulty for every decoder; the fixed-depth architectural gap does not grow monotonically.}
\label{fig:finecops-performance}
\end{figure}

Tuple types add a more specific descriptive view. The largest fixed-depth
deficit appears in the one-hop slice at -1.32 points with a 95\% interval of
[-2.02, -0.49]. The zero-hop deficit is -0.57 points, while the two-hop slice
is -0.08 points with an interval spanning zero. These labels describe the
released construction process, not a complete taxonomy of reasoning. Their
non-monotonic pattern again argues against the simple rule that more hops imply
greater FFN necessity.

\begin{figure}[t]
\centering
\includegraphics[width=0.98\linewidth]{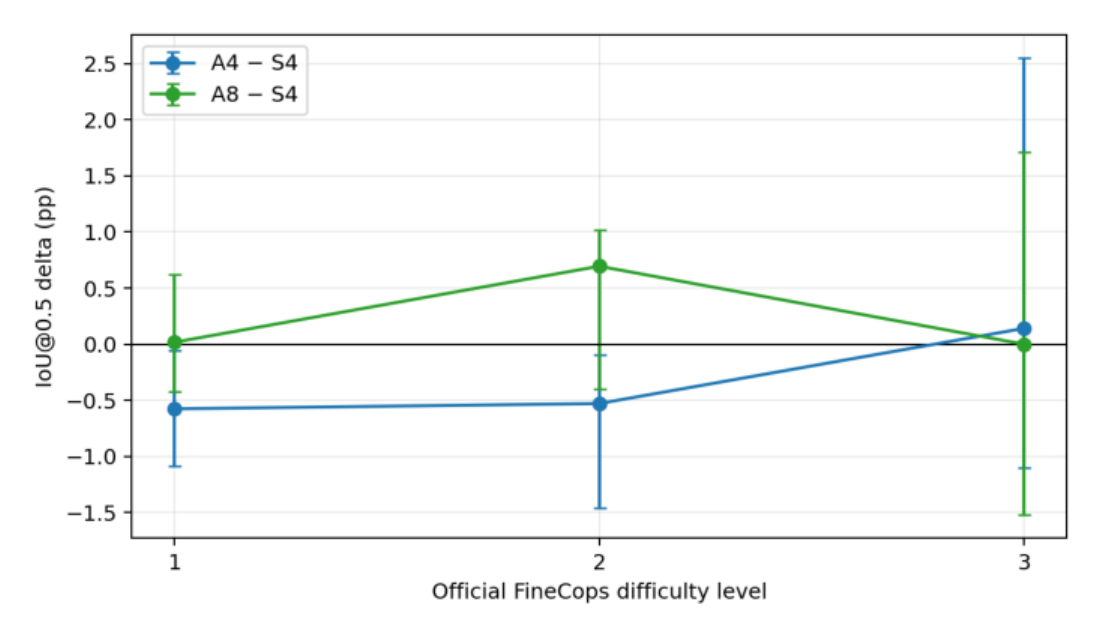}
\caption{FineCops-Ref paired differences by official level and tuple type. The one-hop slice contains the largest observed A4 deficit, but the sequence is not monotonic.}
\label{fig:finecops-deltas}
\end{figure}

\subsection{Does attention-only depth recover the gap?}\label{does-attention-only-depth-recover-the-gap}

A8 reaches 27.79\% on FineCops-Ref, 0.26 points above S4 overall. Its 95\%
interval relative to S4 is [-0.34, +0.52], so the parameter-matched model is
consistent with recovery rather than a new accuracy advantage. At level 2,
A8 is +0.70 points above S4; at level 3, it is effectively identical. The
two-hop tuple slice favors A8 by 0.99 points with an interval above zero, while
the one-hop slice remains -0.23 points with an interval crossing zero.

This pattern changes the causal reading of the A4 deficit. Fixed-depth deletion
does remove useful capacity on FineCops, but a comparable parameter budget
spent on more attention layers restores the overall score. The result is
compatible with attention-only computation performing the needed retrieval and
composition when enough depth is available. It does not prove that A8 learns
the same internal mechanism as S4.

\begin{figure}[t]
\centering
\includegraphics[width=0.98\linewidth]{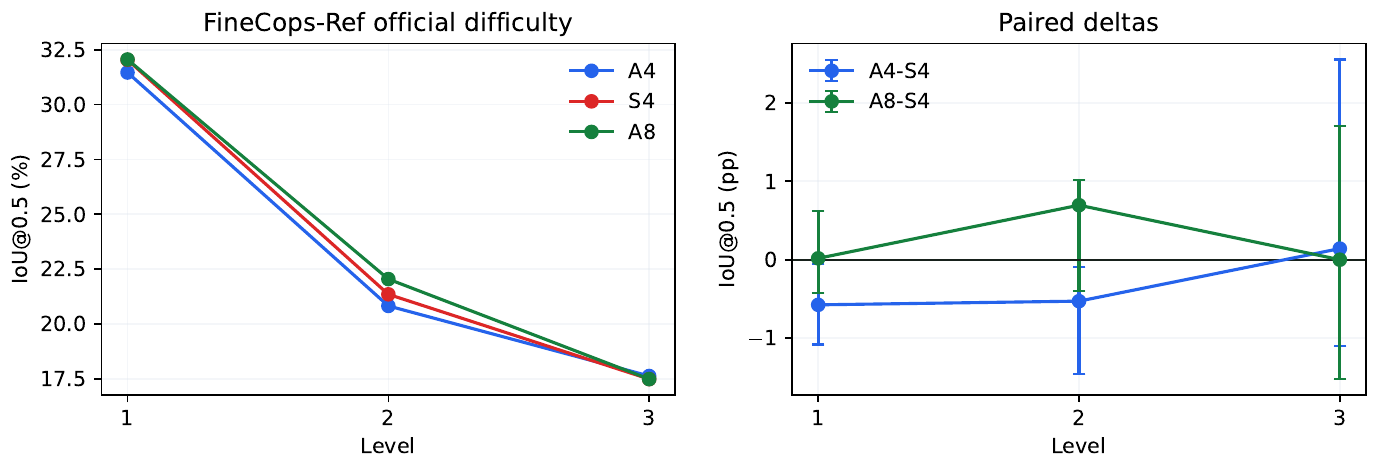}
\caption{FineCops absolute performance and paired intervals together. A8 recovers the small overall fixed-depth gap, while official level 3 remains too noisy for a difficulty-boundary claim.}
\label{fig:finecops-difficulty}
\end{figure}

A8 is also not a compute-matched replacement. Its decoder uses approximately
840 million MACs per example compared with 504 million for S4. The capacity
control answers whether attention depth can substitute under a similar
trainable parameter count. It does not establish the lowest-latency design.

\subsection{Efficiency and second-backbone control}\label{efficiency-and-second-backbone-control}

A4 contains 2,639,104 trainable parameters, 44.4\% fewer than S4's 4,743,424.
Its analytical MAC count is 501.7 million versus 503.8 million for S4 because
cross-attention over the frozen token streams dominates the small FFN
difference. Repeated cached-feature timings are 6.71 ms for A4 and 7.46 ms for
S4, a 10.1\% reduction. Decoder peak memory falls from 817.3 MB to 806.2 MB.

End-to-end timing tells a different story. The complete frozen pipeline takes
209.35 ms for A4 and 210.20 ms for S4, a difference of only 0.4\%. Image and
text preprocessing plus the frozen SigLIP2 backbone dominate total cost. A4 is
therefore smaller and faster as a decoder, but the experiment does not support
a meaningful end-to-end speed claim.

\begin{figure}[H]
\centering
\includegraphics[width=0.98\linewidth]{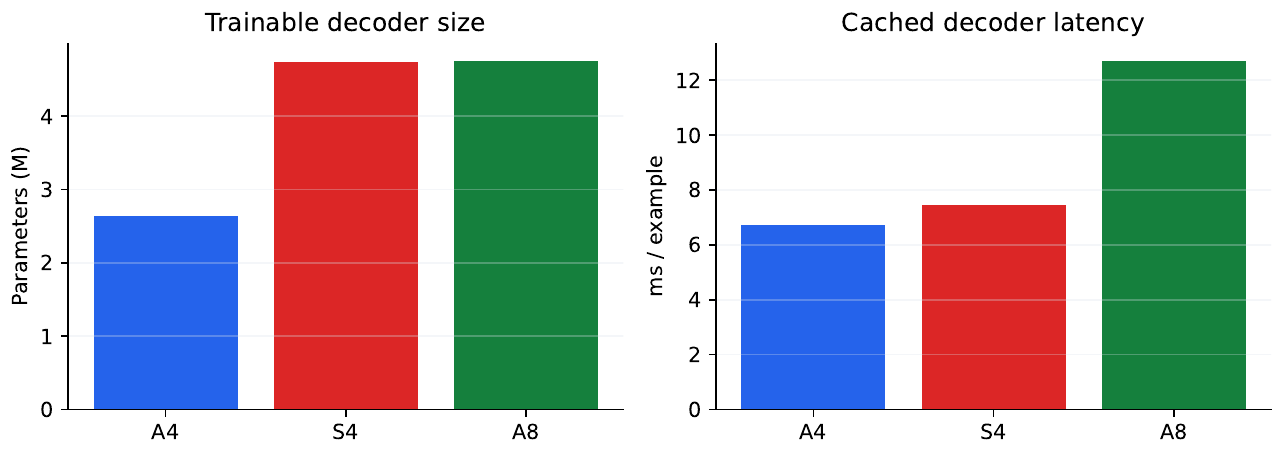}
\caption{Trainable decoder parameters and cached-feature latency. A4 is the efficient fixed-depth variant; A8 is a parameter control and costs more decoder compute.}
\label{fig:efficiency}
\end{figure}

The frozen CLIP control checks direction under a second representation family.
On 9,602 RefCOCOg examples with one seed, A4 reaches 50.89\% and S4 reaches
50.71\% IoU@0.5, a +0.18-point difference. Mean IoU differs by -0.22 points,
while pointing accuracy is identical. The control is consistent with the
primary retention result but remains descriptive because no multi-seed interval
was run.

Together, the experiments support four conclusions. First, fixed-depth FFN
deletion is benign on standard and adversarial grounding. Second, controlled
composition exposes a small but real fixed-depth gap. Third, parameter-matched
attention depth recovers that overall gap. Fourth, neither released difficulty
levels nor adversarial metadata identifies a monotonic region where FFNs become
increasingly necessary.

\section{Limitations}\label{limitations}

The result applies to one small single-query grounding decoder over frozen
features. The backbone itself retains pretrained FFNs and contributes most of
the system's parameters and latency. The paper therefore does not establish a
complete attention-only vision-language model, and it does not support removing
FFNs from the backbone or a generative language decoder.

The output is one box derived from a $24\times24$ patch heatmap. The experiment
does not cover pixel segmentation, multiple objects, open-vocabulary detection,
or autoregressive grounding. Patch resolution and the deterministic mass-to-box
rule limit absolute localization quality, particularly on small objects and
adversarial examples. A higher-resolution or multi-scale readout could change
both absolute scores and the relative value of decoder sublayers.

Only SigLIP2 receives a complete three-seed study. The CLIP experiment is a
one-seed directional control, and RefCOCO is seed-0 descriptive evidence. The
result therefore samples a narrow set of frozen representations and training
runs. No backbone fine-tuning, multiple learned queries, decoder-width sweep,
compute-matched attention control, or FFN hidden-width sweep is included.

The FineCops analysis uses the official positive test set and its released
metadata. Level 3 and several tuple slices are small or imprecise. Ref-Adv-s is
a prepared 1,142-example subset with low absolute accuracy for every variant.
These benchmarks are valuable controlled stresses, but their labels should not
be treated as exhaustive definitions of compositional or adversarial reasoning.

Finally, the paired bootstrap captures uncertainty over observed images after
averaging three seeds. It does not capture uncertainty over backbone pretraining,
dataset families, hyperparameter choices, or alternative decoder scales. A8
recovery shows one viable parameter allocation; it does not identify why every
FFN effect appears or prove equivalence between the learned computations.

\section{Related Work}\label{related-work}

\textbf{Transformer grounding decoders.} MDETR conditions end-to-end detection
on free-form text and jointly reasons over image and language
\citep{kamath2021mdetr}. Grounding DINO combines a feature enhancer,
language-guided query selection, and a cross-modality decoder for open-set
detection and referring-object benchmarks \citep{liu2024groundingdino}. VLT
uses language-conditioned queries and Transformer interactions for referring
segmentation \citep{ding2021vlt}. These systems establish attention-plus-FFN
decoding as a strong standard design. Our work does not replace their full
capabilities; it isolates one FFN ablation in a smaller frozen-feature head.

\textbf{Grounding from frozen representations.} Kang et al. show that a few
internal heads in a frozen LVLM carry strong localization signals
\citep{kang2025heads}. F-LMM derives word-pixel maps from a frozen multimodal
model and retains learned CNN and SAM-based refinement \citep{wu2025flmm}.
ReCLIP repurposes frozen CLIP for proposal scoring and adds a spatial resolver
after showing that off-the-shelf spatial reasoning is weak
\citep{subramanian2022reclip}. These works demonstrate that pretrained features
and attention can support grounding, but they do not compare newly trained
decoders with and without FFNs over identical frozen tokens.

\textbf{Attention-only computation.} The original Transformer pairs
multi-head attention with position-wise FFNs \citep{vaswani2017attention}.
Recent controlled language-model work studies what remains when FFNs are
removed and capacity is reassigned to attention depth
\citep{ndubuaku2026attention}. Sikarwar et al. analyze attention-only
composition in synthetic grounded tasks and show that depth matters for
relational variants \citep{sikarwar2022groundcompose}. Our experiment asks the
corresponding natural-image question under frozen VLM representations and a
single-box output.

\textbf{Referring-expression evaluation.} RefCOCO, RefCOCO+, and RefCOCOg are
standard sources for referring-expression comprehension
\citep{yu2016context}. Ref-Adv introduces adversarial distractors and reasoning
facets designed to suppress common shortcuts \citep{dong2026refadv}.
FineCops-Ref provides controlled compositional annotations and official
difficulty structure \citep{liu2024finecops}. We use these datasets as a
sequence of increasingly controlled stress tests, while keeping the model and
post-processing frozen after RefCOCOg validation.

\section{Conclusion}\label{conclusion}

We asked whether a small visual grounding decoder still needs token-wise FFNs
after a pretrained VLM has supplied contextual image and language features. At
fixed depth, the answer depends on the evaluation. A4 retains standard
RefCOCOg grounding, uses both modalities, and does not collapse on Ref-Adv-s.
FineCops-Ref reveals a small 0.52-point deficit, showing that FFN deletion is not
free in every controlled setting.

The parameter-matched A8 control recovers the FineCops gap. This shifts the
supported conclusion from FFN necessity to capacity allocation: for this
single-query decoder, additional attention depth can substitute for the
removed FFN budget at the measured scale. Official difficulty levels and
adversarial metadata do not reveal a monotonic boundary where attention-only
decoding increasingly fails.

The practical result is bounded. A4 is a smaller and lower-latency decoder, but
the frozen backbone dominates end-to-end cost. The study does not describe a
complete attention-only VLM. It provides controlled evidence that when frozen
representations already expose the relevant visual and linguistic context,
attention-only decoder capacity can often replace FFN capacity for single-box
grounding.

\appendix

\section{Reproducibility Details}\label{reproducibility-details}

The release is evidence-first. \texttt{scripts/generate\_paper\_assets.py}
reads committed bootstrap, slice, efficiency, CLIP-control, and RefCOCOg
summary artifacts. It regenerates the eight manuscript figure families, the
main result table, the machine-readable paper-data manifest, and matching web
assets without a model download or GPU. \texttt{scripts/verify\_paper.py}
checks the frozen threshold, bootstrap seed and replicate count, headline
values, source hashes, expected assets, PDF text, author metadata, and page
count.

\begin{table}[H]
\centering
\small
\setlength{\tabcolsep}{5pt}
\begin{tabular}{p{0.31\linewidth}p{0.61\linewidth}}
\toprule
Artifact & Purpose \\
\midrule
\texttt{configs/study-contract.json} & Frozen model, decoder, metric, and bootstrap contract. \\
\texttt{research-docs/} & Dataset card, final status, and frozen evaluation protocol. \\
\texttt{results/} and \texttt{docs/results/} & Release map, aggregate evidence, per-example metrics where versioned, bootstrap outputs, and plots. \\
\texttt{paper/data/paper-data.json} & Machine-readable source snapshot and SHA-256 input hashes for generated paper artifacts. \\
\texttt{decision-log/} & Compact records of intervention, evaluation, and release decisions. \\
\bottomrule
\end{tabular}
\caption{Primary reproducibility artifacts in the release.}
\label{tab:reproducibility-artifacts}
\end{table}

The repository intentionally excludes third-party images, backbone weights,
provider-local checkpoints, and unversioned qualitative box exports. The paper
therefore does not fabricate qualitative examples from aggregate metrics.
Qualitative panels should be added only from preserved raw predictions and
source-attributed images whose redistribution terms permit publication.

The fixed architecture uses width 256, eight attention heads, four blocks for
A4 and S4, eight blocks for A8, FFN hidden width 1024 for S4, zero dropout, and
one grounding query. The frozen mass threshold is 0.8. The bootstrap uses
10,000 image-clustered replicates at seed 20260812. \texttt{test\_study.py}
checks architecture, masks, target distributions, gradients, and box-conversion
invariants.

The primary backbone revision, dataset preparation decisions, checkpoint
selection boundary, and analysis contract are recorded separately from the
manuscript. This division is deliberate: the paper presents the scientific
argument, while machine-readable contracts and decision logs retain the exact
values needed to audit it. Generated figures are derived from versioned JSON,
CSV, and Markdown evidence rather than values copied manually into plotting
code. Every input file used by the paper-data manifest is stored with its
SHA-256 hash.

To reproduce the CPU-only release artifacts, install the pinned Python paper
requirements, run \texttt{make paper-assets}, and compile \texttt{main.tex}
with Tectonic. \texttt{make verify-paper} repeats generation, compiles the
canonical source, reopens the PDF, and checks expected phrases, metadata, page
count, figure families, and frozen headline values. Model training and dataset
images are not required for this release audit; they are required only to
reproduce the upstream experiments themselves.

\subsection{Frozen evidence snapshot}\label{frozen-evidence-snapshot}

Table \ref{tab:appendix-main-results} collects the comparisons that govern the
paper's conclusion. The intervals retain their original roles: RefCOCOg direct
uses the prespecified 90\% practical-retention interval, Ref-Adv-s and
FineCops-Ref use paired 95\% intervals, and the CLIP control has no multi-seed
interval. The table is generated from the same source artifacts as the figures,
so the manuscript does not maintain a second manually edited results ledger.

\begin{table}[H]
\centering
\small
\caption{Frozen headline comparisons. Positive A4-S4 values favor the attention-only decoder.}
\resizebox{\linewidth}{!}{\input{tables/generated-main-results.tex}}
\label{tab:appendix-main-results}
\end{table}

The decision path is equally important as the point estimates. RefCOCOg passed
direct retention but failed the stronger interaction gate. Ref-Adv-s then
failed to reveal the expected adversarial boundary. FineCops-Ref supplied the
first interval fully favoring S4 at fixed depth, and A8 recovered the overall
score under approximate parameter matching. This sequence was preserved rather
than rewriting the study around only its final positive result.

\begin{table}[H]
\centering
\small
\setlength{\tabcolsep}{5pt}
\begin{tabular}{>{\raggedright\arraybackslash}p{0.22\linewidth}>{\raggedright\arraybackslash}p{0.30\linewidth}>{\raggedright\arraybackslash}p{0.38\linewidth}}
\toprule
Comparison & Question & Permitted reading \\
\midrule
A4 versus S4 & Does fixed-depth FFN deletion change grounding? & Causal decoder comparison with all other trainable components shared. \\
A8 versus S4 & Can attention depth use a similar parameter budget? & Capacity-reallocation evidence; not compute-matched and not mechanistic equivalence. \\
Correct versus shuffle & Does the decoder depend on both modalities? & Diagnostic against the simplest image-only or text-only shortcut. \\
SigLIP2 versus CLIP direction & Does retention depend on one backbone family? & One-seed descriptive transfer only. \\
\bottomrule
\end{tabular}
\caption{Interpretation boundaries for the principal controls.}
\label{tab:interpretation-boundaries}
\end{table}

The released evidence is sufficient to regenerate aggregate plots and verify
the paper's numerical claims without proprietary infrastructure. Full training
reproduction remains more demanding: it requires the pinned frozen backbones,
upstream datasets under their own terms, cached or uncached feature extraction,
and the fixed training runs for every seed. Provider-local checkpoints and raw
predictions that were not versioned are not silently reconstructed from summary
tables. Where those artifacts are absent, the release states the boundary
instead of presenting an unverifiable qualitative result.

This audit structure also separates scientific revisions from formatting
changes. A later manuscript version may rearrange prose or figures while the
machine-readable contract continues to reject altered thresholds, bootstrap
settings, headline values, or input hashes. New experiments should enter as new
versioned evidence with an explicit decision record; they should not overwrite
the frozen first-manuscript snapshot.

\subsection{Released difficulty values}\label{released-difficulty-values}

Tables \ref{tab:finecops-slices} and \ref{tab:refadv-slices} preserve the slice
values behind the difficulty figures. They are included to make the
non-monotonic conclusion inspectable without estimating values from plots.
FineCops levels are official. Ref-Adv quartiles are inclusive empirical bins
over released metadata, fixed before decoder-performance differences were
examined.

\begin{table}[H]
\centering
\small
\begin{tabular}{lrrrrl}
\toprule
FineCops slice & N & A4 & S4 & A8 & A4-S4 95\% CI \\
\midrule
Level 1 & 5,730 & 31.47 & 32.04 & 32.06 & [-1.08, -0.06] \\
Level 2 & 3,404 & 20.82 & 21.35 & 22.04 & [-1.46, -0.09] \\
Level 3 & 471 & 17.62 & 17.48 & 17.48 & [-1.10, +2.55] \\
Zero hop & 2,333 & 31.22 & 31.79 & 32.30 & [-1.33, -0.01] \\
One hop & 2,146 & 26.72 & 28.04 & 27.80 & [-2.02, -0.49] \\
Two hop & 2,555 & 21.97 & 22.05 & 23.04 & [-0.65, +0.87] \\
\bottomrule
\end{tabular}
\caption{FineCops-Ref IoU@0.5 percentages by selected official slices.}
\label{tab:finecops-slices}
\end{table}

The FineCops values show why the overall deficit and the boundary question must
remain separate. Levels 1 and 2 favor S4 by similar small amounts, while level
3 reverses sign with much greater uncertainty. The one-hop tuple slice is the
largest fixed-depth deficit, but the two-hop slice is near zero and A8 exceeds
S4 there. Neither ordering supports a simple more-composition-means-more-FFN
rule.

\begin{table}[H]
\centering
\small
\begin{tabular}{lrrrrl}
\toprule
Ref-Adv slice & N & A4 & S4 & A8 & A4-S4 95\% CI \\
\midrule
Negation & 457 & 7.29 & 7.44 & 8.10 & [-1.24, +1.02] \\
No negation & 685 & 8.66 & 7.25 & 7.79 & [+0.49, +2.34] \\
Longest quartile & 261 & 8.81 & 7.02 & 7.92 & [+0.26, +3.45] \\
Most distractors & 191 & 4.89 & 4.89 & 4.71 & [-1.05, +1.05] \\
\bottomrule
\end{tabular}
\caption{Ref-Adv-s IoU@0.5 percentages on selected native-metadata slices.}
\label{tab:refadv-slices}
\end{table}

Ref-Adv provides the complementary negative result. Negation is inconclusive,
the longest-expression quartile favors A4, and the highest-distractor quartile
is tied at fixed depth. These slices do not identify a region where S4's FFNs
become increasingly advantageous. Their low absolute scores remain visible in
the table and prevent a paired retention result from being mistaken for strong
adversarial performance.

\section*{Acknowledgements}

AI-assisted tools were used for code assistance, experiment orchestration,
figure generation, manuscript organization, and language editing. No AI system
is an author; responsibility for the submitted work remains with the human
author.

\clearpage
\bibliography{references.bib}

\end{document}

%% file: tables/generated-main-results.tex
\begin{tabular}{p{0.27\linewidth}rll}
\toprule
Evaluation & A4--S4 (pp) & Interval & Reading \\
\midrule
RefCOCOg direct (3 seeds) & +0.26 & 90\% CI [-0.33, +0.84] & Direct-retention gate passed; interaction not confirmed \\
Ref-Adv-s overall (3 seeds) & +0.79 & 95\% CI [+0.06, +1.52] & No monotonic hard-slice boundary \\
FineCops-Ref overall (3 seeds) & -0.52 & 95\% CI [-0.95, -0.12] & A8-S4 = +0.26 pp \\
CLIP RefCOCOg (1 seed) & +0.18 & descriptive only & Second frozen backbone \\
\bottomrule
\end{tabular}